\documentclass[conference]{IEEEtran}
\IEEEoverridecommandlockouts
\usepackage{cite}
\usepackage{amsmath,amssymb,amsfonts}
\usepackage{algorithmic}
\usepackage{graphicx}
\usepackage{textcomp}
\usepackage{xcolor}

\usepackage{algorithm}
\usepackage{algorithmic}
\usepackage{amsthm,amsmath,amssymb}
\usepackage{enumitem} %
\usepackage{colortbl}  %
\usepackage{soul} %
\usepackage{color, xcolor} 
\usepackage{tabularx}
\usepackage[most]{tcolorbox}
\usepackage{multirow}
\usepackage{booktabs}
\usepackage{pifont}
\usepackage{array}

\newcommand{\ourmethod}{\text{DyRate}}

\newcommand{\ourmethodName}{\textbf{Dy}namic \textbf{Rate} (\textbf{DyRate})}
\usepackage{newfloat}
\usepackage{listings}
\usepackage{subfigure}
\usepackage{makecell}
\newcolumntype{L}[1]{>{\raggedright\arraybackslash}m{#1}}  
\newcolumntype{C}[1]{>{\centering\arraybackslash}m{#1}}  

\def\BibTeX{{\rm B\kern-.05em{\sc i\kern-.025em b}\kern-.08em
    T\kern-.1667em\lower.7ex\hbox{E}\kern-.125emX}}

\usepackage{authblk}
\title{Dynamic Token Reduction during Generation for Vision Language Models}

\author[1]{Xiaoyu Liang\thanks{$^{*}$These authors contributed equally to this work.}$^{*}$}
\author[1]{Chaofeng Guan$^{*}$}
\author[1]{Jiaying Lu}
\author[2]{Huiyao Chen}
\author[3]{Huan Wang}
\author[1]{Haoji Hu\thanks{$^{\dagger}$Corresponding author.}$^{\dagger}$}

\affil[1]{Zhejiang University, Hangzhou, China}
\affil[2]{Harbin Institute of Technology, Shenzhen, China}
\affil[3]{Westlake University, Hangzhou, China}

\begin{document}

\maketitle

\begin{abstract}
Vision-Language Models (VLMs) have achieved notable success in multimodal tasks but face practical limitations due to the quadratic complexity of decoder attention mechanisms and autoregressive generation. Existing methods like FASTV~\cite{FastV} and VTW~\cite{VTW} have achieved notable results in reducing redundant visual tokens, but these approaches focus on pruning tokens in a single forward pass without systematically analyzing the redundancy of visual tokens throughout the entire generation process. 
In this paper, we introduce a dynamic pruning strategy tailored for VLMs, named \ourmethodName{}, which progressively adjusts the compression rate during generation.  
Our analysis of the distribution of attention reveals that the importance of visual tokens decreases throughout the generation process, inspiring us to adopt a more aggressive compression rate. 
By integrating a lightweight predictor based on attention distribution, our approach enables flexible adjustment of pruning rates based on the attention distribution.  
Our experimental results demonstrate that our method not only reduces computational demands but also maintains the quality of responses.
\end{abstract}


\section{Introduction}
\label{sec:intr}

Vision-Language Models (VLMs) integrate visual and textual information to generate coherent and contextually relevant outputs, demonstrating impressive capabilities across various tasks in artificial intelligence~\cite{bai2023qwen,lin2023video,liu2023llava}.
However,  the autoregressive generation nature of VLMs, along with the quadratic complexity of attention~\cite{SurveyonLLM}, significantly increases computational complexity when the input token sequence lengthens, limiting the applications of VLMs.

Many recent methods attempt to reduce the computational cost by eliminating redundant visual tokens. 
For example, FastV~\cite{FastV} analyzes the attention mechanism and discovers that visual tokens receive less attention after the second decoder layers, thus deleting redundant visual tokens according to a predefined compression rate. 
Building on this observation, VTW~\cite{VTW} directly removes all visual tokens in deeper layers, only text tokens allowed to participate in further processing.

\begin{figure}[htbp]
	\includegraphics[width=\columnwidth]{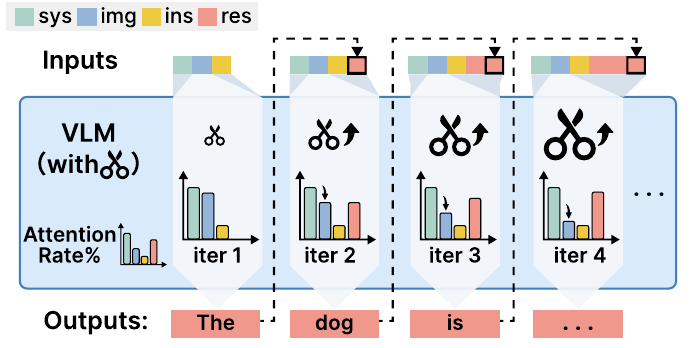}
	\caption{As the VLMs continue to generate, the attention proportions among the four token types (system, image, instruction, response) fluctuate.
Notably, as the number of iterations increases, visual tokens receive decreasing attention. This observation suggests that we can adjust the appropriate compression rate $R$ based on the attention distribution to implement more aggressive pruning during the model's generation process.}
	\label{fig:first}
\end{figure}

Although these methods demonstrate effectiveness, they still present two drawbacks. 
\textit{First}, the compression rate must be manually specified. 
Manually selecting an appropriate compression rate is nontrivial and typically requires expert-level domain knowledge. 
\textit{Second}, maintaining a fixed compression rate during the generation process is suboptimal.  
Our experimental analyses of the attention distribution during generation (see Sec.~\ref{sec:vr}) reveal that visual tokens receive varying proportions of attention during the next token prediction. 
Information in VLMs often becomes less concentrated in visual tokens during the later iteration of generation. As the generation process advances and the length of the response gradually increases, the importance of the visual information changes accordingly. %
Thus, it is imperative to have \textit{adaptive} compression rates during generation. Yet, determining proper compression rates is challenging. Our work is here to address the challenge.

We present \ourmethodName{}, the \textit{first} method to connect the token reduction rate with the attention weights to achieve adaptive dynamic reduction during the VLM generation process. 
Our approach is designed to be \textit{differentiable}, enabling end-to-end training. 
Specifically, we introduce a lightweight classifier to predict the reduction rate. During each iteration of model generation, we collect the attention distributions of four types of tokens in each head, which are then fed into the predictor to determine the optimal reduction rate for cropping the image embeddings. The overview of our approach is shown in Fig.~\ref{fig:first}.
Our contributions are summarized as follows:

\begin{itemize}[leftmargin=*,nosep]
	\item[$\star$] %
	we propose \ourmethod{}, the \textit{first} approach that can adaptively adjust the token reduction rate during model generation.

	\item[$\star$] %
	we present an effective training strategy based on Gumbel-Softmax  which enables end-to-end training of our proposed module;
	
	\item[$\star$] %
	through extensive experiments, we demonstrate the effectiveness of our approach in maintaining accuracy while reducing computational demands. 
\end{itemize}

\section{Related Work }
\subsection{Efficient VLMs} 
The computational demands of VLMs increase significantly with long token sequences due to the quadratic complexity of attention mechanisms. This limitation restricts their application in tasks such as high-resolution image understanding~\cite{BeyondLLaVA-HD} and video understanding~\cite{Video-llama}. To address this issue, prior research has proposed several strategies. For instance, MoE-Llava~\cite{MoE-LLaVA} integrates a Mixture of Experts framework to accelerate the model, while VL-Mamba~\cite{VL-Mamba} explore alternative architectures to enhance efficiency. Another approach involves using smaller language models, such as LLaVA-Phi~\cite{LLaVA-Phi} and mobileVLM~\cite{MobileVLM}, %
which demonstrate efficiency with minimal performance loss. Additionally, compression techniques like pruning~\cite{LLM-Pruner}, quantization~\cite{LLM.int8(),GPTQ}, and knowledge distillation~\cite{Self-Distillation} are widely used to reduce the number of parameters in models.

However, these methods often require modifications to the model architecture or parameters, complicating further development. In contrast, token reduction minimizes token sequences without altering the model architecture, addressing the quadratic complexity of VLMs. Token reduction has been extensively explored and validated within encoder architectures like ViT and BERT, primarily through pruning~\cite{DynamicViT,EViT,LTP} and merging~\cite{DiffRate,Tom,GroupViT} strategies.

\subsection{Token Reduction for VLMs} 

Recent researchers have applied token reduction methods to VLMs, categorizing them based on the architecture of VLMs. For the visual encoder, methods like LLaVA-PruMerge~\cite{LLaVA-PruMerge} and MADTP~\cite{MADTP} introduce adaptive methods to reduce visual tokens, significantly decreasing their number while maintaining comparable performance to the original models. 
In the cross-modality projector, Tokenpacker~\cite{tokenpacker} optimizes the bridge between textual and visual information to minimize the number of visual tokens. FastV~\cite{FastV}, Sparsevlm~\cite{zhang2024sparsevlm} and Visionzip~\cite{yang2024visionzip} removes redundant visual tokens during the inference phase to reduce computational demands without impacting performance. VTW~\cite{VTW} observes that visual tokens are not significant in deeper layers of the VLM and strategically removes all visual tokens from specific layers, allowing only text tokens to proceed in subsequent processing.

Although existing token reduction methods improve the efficiency of inference, they often overlook the autoregressive nature of VLM decoders. Our work addresses this gap by dynamically adjusting \( R \) based on the changing attention distribution during the generation process.

\section{Methods}
\begin{figure}[!htbp]
	\includegraphics[width=1.0\linewidth]{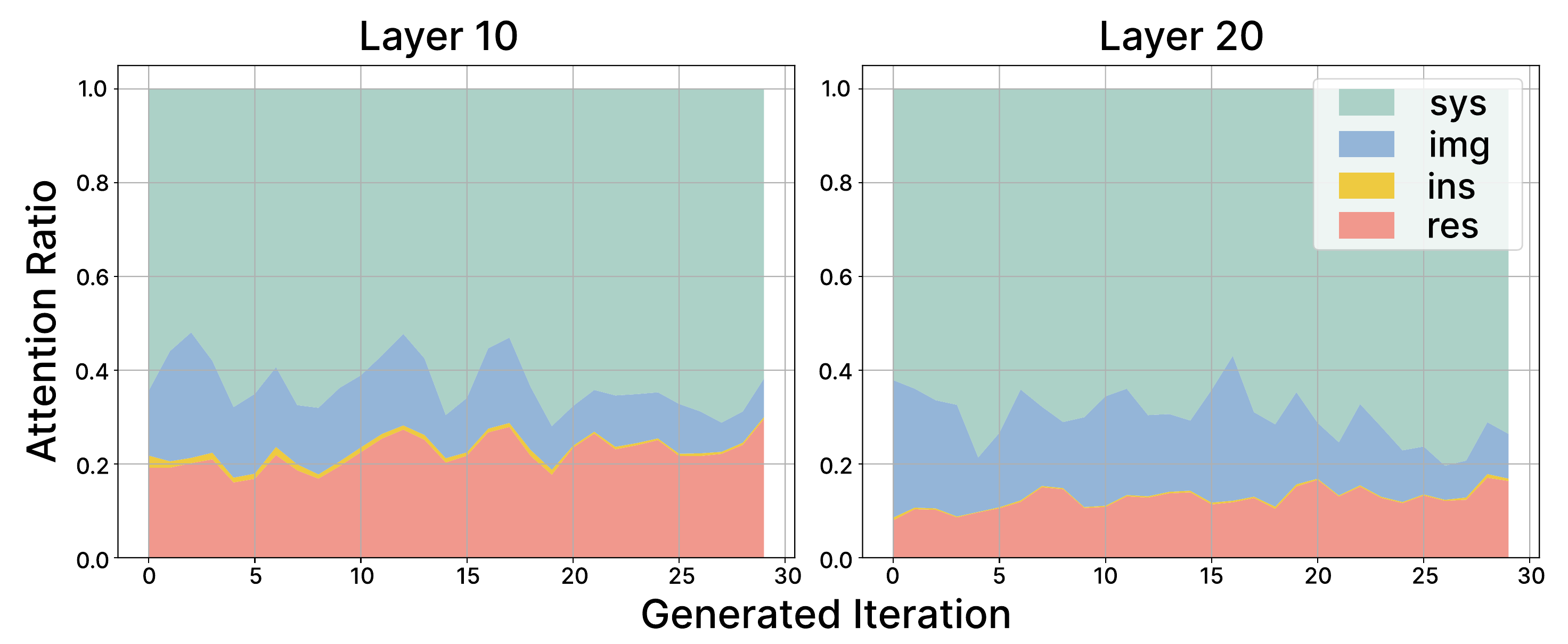}
	\caption{\textbf{Attention stack per iteration} during 
		the decoding process of LLaVAv1.5-7B on the Flickr30K~\cite{flickr30k} dataset, the x-axis represents the time step in generations.
		The left graph depicts shallow layers, while the right graph represents deep layers. Our findings indicate that as generations progress, the importance of visual tokens gradually decreases.
            We categorize the input tokens at each iteration into four types: system prompt (sys), image token (img), user instruction (ins), and response token (res). 
	}
	\label{fig:atten}
\end{figure}

In Sec.~\ref{sec:vr}, we explore how attention distribution changes during the generation within VLMs. Our analysis reveals that the importance of visual tokens gradually decreases as the VLM progresses. 
This observation leads us to question whether the fixed compression rate \(R\) set manually in previous works~\cite{FastV,VTW} is the optimal solution, considering the changing attention distribution.

\begin{figure*}[htbp]
	\centering
	\includegraphics[width=1.0\linewidth]
	{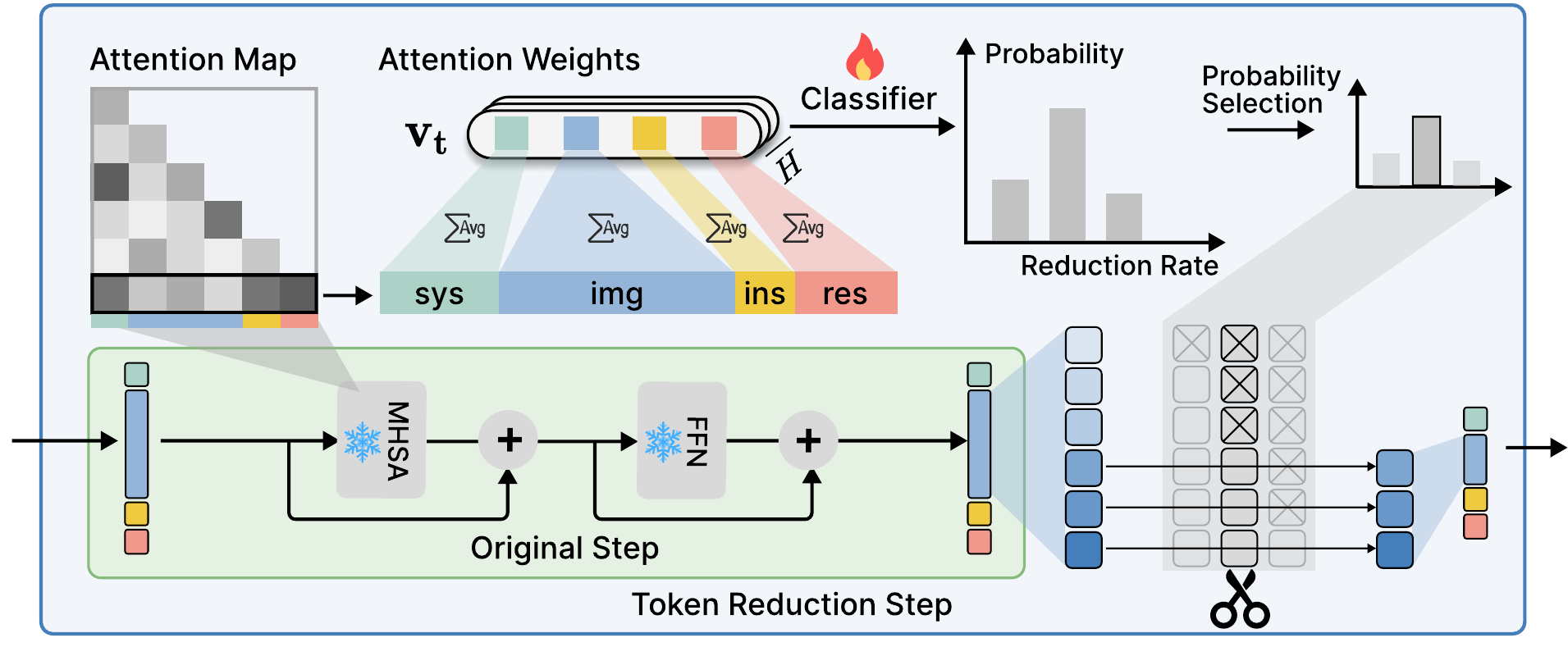}
	\caption{The pipeline of our method. We calculate the attention distribution for each attentional head and train a linear classifier to find the optimal token pruning rate \( R \). The classifier can be trained end-to-end.
    }
	\label{fig:overall}
\end{figure*}

Based on this insight, we propose \ourmethod{} as a solution to adjust the compression rate $R$. %
The overall process is illustrated in Fig.~\ref{fig:overall}.

\subsection{Visual Redundancy during Generation}
\label{sec:vr}

We observed that visual token redundancy increases with generation steps.
In Fig.~\ref{fig:atten}, as the model generates more tokens, it emphasizes response tokens and overlooks visual tokens, indicating increasing redundancy.

According to information flow theory~\cite{infoflow}, a large amount of information from image tokens is aggregated into respones tokens during generation. The aggregation of this information results in additional redundancy among visual tokens, where a large number of tokens provide minimal information, leading to a waste of computational resources. 

Given the above, a fixed \( R \) fails to adapt to the changing attention distribution during the generation process, highlighting the need to determine an adaptive \( R \).

\subsection{Token Reduction with Prediction Module}
\label{sec:diff}

To address this issue, we first intuitively hypothesize that the redundancy of visual tokens is closely related to the attention distribution across four different token types. At each time step of the model generation, we compute the attention distribution for each attention head. Based on these distribution characteristics, we further train a linear classifier aimed at identifying and determining the optimal token pruning rate \( R \).
The entire process is depicted in Figure~\ref{fig:overall}.

During training, we use the mask \(M\) to selectively prune tokens, while during inference, we discard tokens directly to save computational resources. 
It's worth to note that the FLOPs required by the predictor is minimal when compared to the computational savings gained from discarding these tokens. 

\subsection{Differentiable Compression Rate}

To effectively address the non-differentiability issues associated with sampling and masking operations, 
we utilize the Gumbel-Softmax trick~\cite{DynamicViT,DiffRate,chen-etal-2024-semantic} to convert the probability distribution of the compression rate \( R \) predicted by the predictor \( C \) into a differentiable mask probability distribution, which is then integrated into the forward propagation of the model.

\subsubsection{Probability of Pruning Rate}%

We define the adjustment of the compression rate \(R\) as a classification problem. 
and discretize the compression rate \( R \) into \( K \) discrete values:
\begin{equation}
	\mathcal{R} = \{ r_k \}_{k=1}^{K},
\end{equation}
where \( r_k = \frac{k-1}{K} \) represents removing \(\frac{(k-1) \cdot N}{K}\) less important tokens in a token sequence of length \( N \). 
Note that \( r_1 = 0 \) indicates retaining all tokens.

We feed the vector \( \mathbf{v}_t \) to classifier \( C \) to predict the probability distribution over these rates:
\begin{equation}
	\label{eqn:softmax}
	\pi_R = Softmax(C_\theta(\mathbf{v}_t))= \{ P(r_k) \}_{k=1}^{K},
\end{equation} 
where the probabilities sum to one and are differentiable.

Specifically, we sort the tokens based on the `attenion-score' rule utilized in FastV~\cite{FastV}. 
After that, we apply masks at different pruning rates:
\begin{equation}
	m_i^k =
	\begin{cases}
		1 & \text{if } i \leq \frac{N}{K}(k-1), \\
		0 & \text{otherwise},
	\end{cases}
\end{equation}
where 0 indicates an unimportant token to be discarded, and 1 indicates a token to be retained. For each potential \(R_i\), we generate a unique corresponding mask \(M_i\), ensuring that the least important tokens are discarded first.

\subsubsection{Gumbel-Softmax Sampling}
During the forward pass, Gumbel-Softmax generates a one-hot vector with the same expectation as \(\pi_{R}\):
\begin{equation}
	R = \text{Gumbel-Softmax}(\pi_R) \in \left \{ 0,1 \right \} ^K,
\end{equation}
where their parameter gradients can be easily computed with standard backpropagation. %
During the backward pass, the straight-through Gumbel-Softmax estimator is employed to approximate the gradient.

We can then select current mask M by sampling from $\pi$:
\begin{equation}
	m = \sum_{k=1}^K \text{Gumbel-Softmax}(\pi_R)_{*,k} \cdot m^k ,
\end{equation}%
where \(m_k\in \left \{ 0,1 \right \}^N\) is the mask associated with the discrete compression rate \(r_k\).

This generates the final mask \(M\in \left \{ 0,1 \right \}^{N\times N}\), which, in conjunction with the causal mask, is applied to the attention computation to achieve token pruning.

\begin{equation}
	\label{eq:masking_operator}
	M_{i,j} =
	\begin{cases}
		1~~&i=j,\\
		m_i~~&i\ne j.
	\end{cases}
\end{equation}

The process ensures that the selection operation remains differentiable with respect to \( R \).
And we keep the proof and pseudocode for our \ourmethod{} algorithm in the Appendix.

\section{Experiments}

\begin{table*}[!t]
\caption{\textbf{Comparison among different VLMs on 4 visual question answering benchmarks and 3 common benchmarks.}
Benchmark names are abbreviated due to space limits. The highest-performing results are highlighted in \textbf{boldface}. }
\label{tab:tongyong}
\small
\centering
\begin{tabular}{lcc|cccc|ccc}
\toprule
\multirow{2}{*}{\textbf{Model}} & \multirow{2}{*}{\textbf{LLM}}  & \multirow{2}{*}{\textbf{Res.}}&  \multicolumn{4}{c|}{\textbf{Visual Question Generation}} & \multicolumn{3}{c}{\textbf{Short Generation}} \\
&  & &  GQA↑ & VisWiz↑ & SQA$^\text{I}$↑ & VQA$^\text{T}$↑ & POPE↑  & MMB↑ & MME↑  \\
&  & &  \textit{EM}& \textit{EM} & \textit{EM} & \textit{EM} & \textit{Accuracy} & \textit{Accuracy} & \textit{Accuracy} \\
\midrule

InstructBLIP & Vicuna-7B & 224 & 49.2 & 34.5 & 60.5 & 50.1 & 79.8  & 36.0 & -- \\
InstructBLIP & Vicuna-13B & 224 & 49.5 & 33.4 & 63.1 & 50.7 & 78.9 & -- & 1212.8 \\
MiniGPT-4 & Vicuna-13B & 224 & 41.0 & 19.6 & 61.0 & 42.5 & 85.3  & -- & 1293.8 \\
Qwen-VL & Qwen-7B & 448& 59.3 & 35.2 & 67.1 & {63.8} & --  & 38.2 & -- \\
Qwen-VL-Chat & Qwen-7B & 448 & 57.5 & 38.9 & 68.2 & 61.5 & --  & 60.6 & 1487.5 \\

\midrule
LLaMA-VID & Vicuna-7B & 336 & 64.3 & 54.1 & 68.3 & -- & 86.0  & 63.4 & 1521.4 \\
VoCo-LLaMA & Vicuna-7B& 33 & 57.0 & 53.0 & 65.4 & 52.7 & 81.4  & 58.8 & 1323.3 \\
TokenPacker & Vicuna-7B & 336 & 61.9 & {52.0} & {--} & -- & 87  & 65.1 & -- \\
M$^3$ & Vicuna-7B & 336 & 61.3 & 53.1 & 67.2 & -- & 86.6  & 63.6 & -- \\
PruMerge & Vicuna-7B & 336 & 61.8 & 53.5 & {68.5} & 56.0 & 76.3  & 60.9 & 1350.3 \\
LLaVAv1.5 & Vicuna-7B & 336 & 62.0 & {50.0} & {66.8} & 58.2 & 85.9  & 64.3 & 1510.7 \\
\midrule
FastV & Vicuna-7B & 336 & 60.3  & \textbf{54.4}  & 69.0  & 45.4  &  82.5 &  63.9  & 1510.2 \\
VTW & Vicuna-7B & 336 & 55.1  & 50.9  & 69.1  & 16.1  &  85.9 & 64.0  & 1501.4 \\
SparseVLM & Vicuna-7B & 336 & 57.6 &--& 69.1&56.1&83.6& 62.5& 1721 \\
VisionZip & Vicuna-7B & 336 &60.1&--& 68.9&57./&84.93&63.4&\textbf{1834}\\

\rowcolor{blue!15}
\ourmethod(ours) & Vicuna-7B & 336 & \textbf{61.9}  & 54.2  & \textbf{69.2}  & \textbf{45.7}  &  \textbf{86.8} & \textbf{64.1}  & 1516.6 \\         
\bottomrule
\end{tabular}
\end{table*}

\subsection{Settings}
We utilized LLaVA-1.5-7B and LLaVA-1.5-13B models, where visual inputs were represented using 576 tokens. 
To evaluate the general performance of our proposed method, we conducted comparisons with existing models and mainstream token compression techniques across seven widely-used benchmarks. Additionally, we employed three long-response tasks to simulate its capabilities in daily conversation.

We conducted systematic evaluations of the model's performance across multiple datasets using the LMM-Eval evaluation framework on eight NVIDIA GeForce RTX 4090 GPUs.
More details in Appendix.

\subsection{Main Results}

\begin{table}[!t]
\caption{CIDEr Scores of Different Methods on Nocaps, Flickr30K, and COCO2017 Datasets.}
\label{tab:fastv+vtw+back_part2}
\centering
\begin{tabular}{ll|ccc}
\toprule
\multirow{2}{*}{\textbf{Models}} & \multirow{2}{*}{\textbf{Methods}}& Nocaps↑ & Flickr30k↑ & COCO2017↑ \\
&& \textit{CIDEr} & \textit{CIDEr} & \textit{CIDEr}\\
\midrule
\multirow{6}{*}{\shortstack{LLaVA-1.5\\-7B}} 
& Original & 74.89 &105.57&110.43 \\
& FastV$_\text{(K=3,R=0.5)}$ &74.75 &105.00&110.80\\
& FastV$_\text{(K=2,R=0.5)}$ &74.86&104.00&110.40\\
& VTW$_\text{(K=16,R=1)}$ &44.54 &58.00&67.20 \\
\rowcolor{blue!15}
&\ourmethod($ours$) &75.00&108.41  &110.54 \\
\midrule
\multirow{6}{*}{\shortstack{LLaVA-1.5\\-13B}} 
& Original & 109.31&79.56 & 115.57 \\
& FastV$_\text{(K=3,R=0.5)}$ &102.70&73.40 &105.36 \\
& FastV$_\text{(K=2,R=0.5)}$ &103.10&73.40&108.85 \\
& VTW$_\text{(K=16,R=1)}$ &95.21&65.87 &101.67\\
\rowcolor{blue!15}
&\ourmethod(ours) &113.23 &79.38  &115.72  \\
\bottomrule
\end{tabular}
\end{table}

\subsubsection{Short Responses}

Our model demonstrated exceptional performance on short-response datasets, showcasing superior question comprehension and answer generation capabilities. It performed particularly well on the Flickr30K and GQA datasets and handled specific query types well on the SQA[I] and VisWiz datasets. Notably, performance remained stable across different datasets, even under high compression rates, although a minor performance drop was observed during detailed visual analysis tasks.

\subsubsection{Long Responses}

With long-response datasets, our model's strategy of gradually adjusting the pruning ratio during autoregressive generation effectively reduced redundancy. Despite reducing visual tokens by half, the model maintained high performance across multiple datasets. This strategy proved especially beneficial for generating long responses and managing high redundancy data, all while requiring less computational resources.

\subsection{Efficiency}

We are also concerned with the performance of our method in scenarios involving long responses, as such situations frequently arise in real-life applications. 
Therefore, we selected the Flickr30K, COCO2017 and Nocaps dataset for evaluation.
These datasets feature response token lengths ranging from 16 to 64 tokens, providing a comprehensive basis for evaluating the comprehension abilities of VLMs and the effects of post-token pruning.
For a fair comparison, we evaluated the performance of two highly relevant methods, FastV and VTW under scenarios with FLOPs savings ratio of $30\%$ and $50\%$.

The experimental results demonstrate that our method, Diffrate, effectively meets FLOPs resource constraints by automatically optimizing the parameter R, eliminating the need for manual searching. This adaptive approach to R proves to be more effective. For ease of comparison with FastV, we selected the same number of layers.

\begin{figure*}[!htbp]
    \centering
    \includegraphics[width=0.8\linewidth]
    {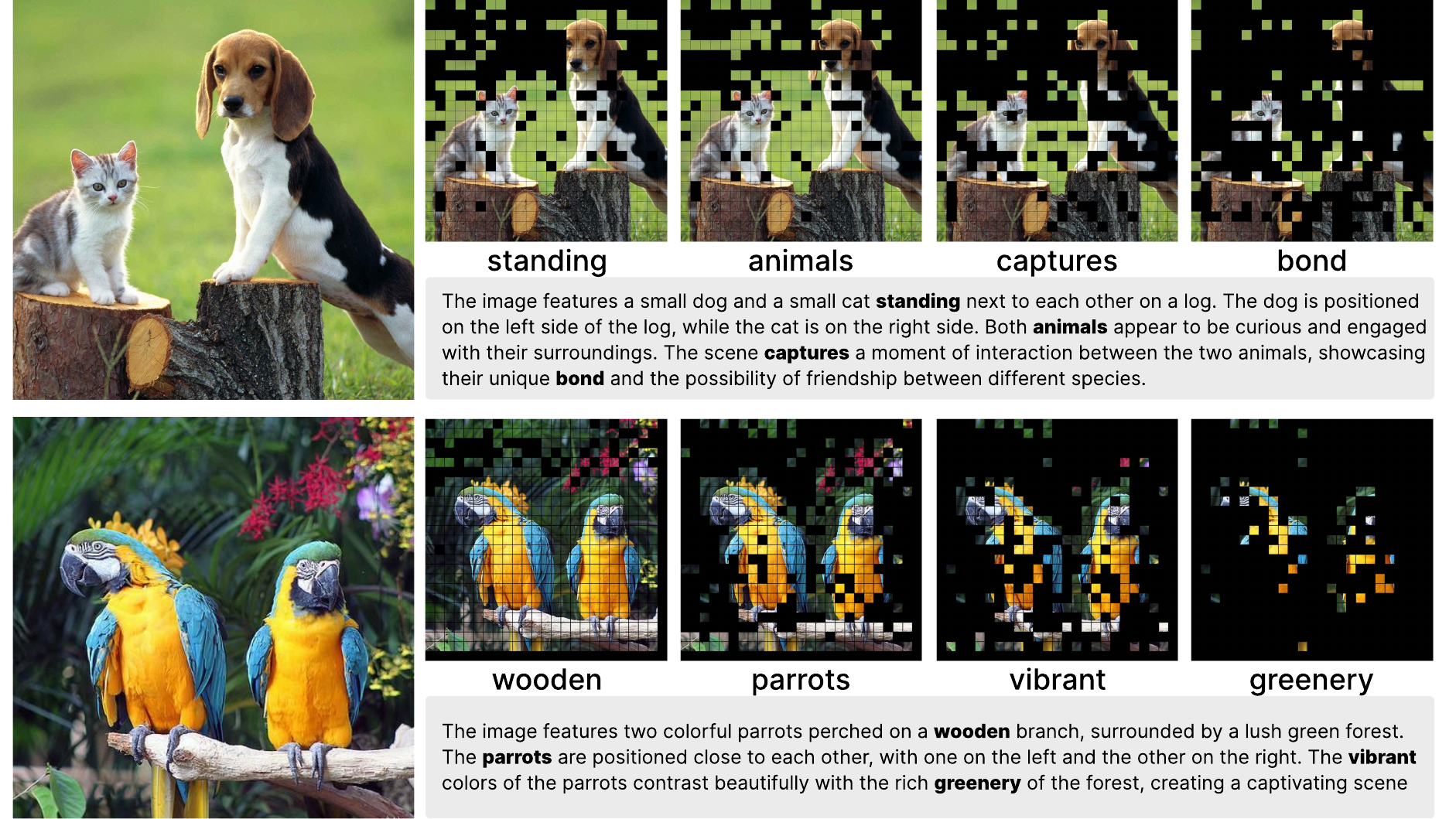}
    \caption{We explore the mask of the image for each token iteration in the model generation process.}
    \label{fig:viz2}
\end{figure*}

\begin{table}[!t]
\caption{\textbf{Comparing Token Reduction Methods for Complex scene description.}}
\label{tab:fastv+vtw+back_part1}
\centering
\resizebox{\columnwidth}{!}{ 
\begin{tabular}{l|cccc}
\toprule
\textbf{Methods} & \textbf{TFLOPs$_{(\%)}$}↓ & \textbf{Latency (ms)}↓ & \textbf{Nocaps}↑ & \textbf{Auto R}\\ 
\midrule
\multicolumn{5}{l}{\textbf{LLaVA-1.5-7B}} \\
\midrule
Original & 100.0 & 70.80 & 74.89 & \ding{55}\\
FastV$_\text{(K=3,R=0.5)}$ & 57.90 & 42.36 & 74.75 & \ding{55}\\
FastV$_\text{(K=2,R=0.5)}$ & 55.40 & 41.37 & 74.86 & \ding{55}\\
VTW$_\text{(K=16,R=1)}$ & 55.20 & 46.24 & 44.54 & \ding{55}\\
\rowcolor{blue!15}
\ourmethod(ours) & 33.33 & 40.13 & 75.00 & \checkmark\\
\midrule
\multicolumn{5}{l}{\textbf{LLaVA-1.5-13B}} \\
\midrule
Original & 100.0 & 128.36 & 109.31 & \ding{55}\\
FastV$_\text{(K=3,R=0.5)}$ & 57.90 & 73.35 & 102.70 & \ding{55}\\
FastV$_\text{(K=2,R=0.5)}$ & 55.40 & 72.25 & 103.10 & \ding{55}\\
VTW$_\text{(K=16,R=1)}$ & 55.20 & 80.78 & 95.21 & \ding{55}\\
\rowcolor{blue!15}
\ourmethod(ours) & 33.33 & 65.16 & 113.23 & \checkmark\\
\bottomrule
\end{tabular}
}
\end{table}

\begin{table}[!t]
\caption{\textbf{Ablation Experiments for LLaVAv1.5-7B on Nocaps dataset.} }
\label{tab:ablation}
\begin{center}
\subfigure[\textbf{Generation Type}\label{tab:ablation_generate}]{
\begin{tabular}{L{1cm}C{1cm}}
\toprule
Strategy&\makecell[c]{Nocaps↑\\ \textit{CIDEr}}\\
\midrule
Greedy & 106.74\\
Beam-2 & 109.06\\
Beam-5 & 108.28\\
top\_p & \textbf{109.57}\\
\bottomrule
\end{tabular}}
\hspace{0.01\linewidth}
\subfigure[\textbf{Pruning Strategy}\label{tab:ablation_textual}]{
\begin{tabular}{L{1.5cm}C{1cm}C{1cm}}
\toprule
Strategy&\makecell[c]{Nocaps↑\\\textit{CIDEr}}&\makecell[c]{FLOPs↓\\(\%)}\\
\midrule
LLaVA & 105.57 & 100.00\\
FastV & 105.00 & 57.90 \\
Ours($FP$) & 107.00 & 28.80 \\
Ours($DP$) & 105.00 & 58.10\\
\bottomrule 
\end{tabular}}
\end{center}
\end{table}

\subsection{Ablation}

\subsubsection{Compression Rate Strategies}

Our ablation studies explored the effects of different pruning strategies on Visual Language Models (VLM) performance. We compared FixedPrune(FP), a strategy with a static pruning ratio, and DepthBasedPrune(DP), a strategy with a dynamically adjusted pruning ratio based on layer depth. These strategies were evaluated on the NoCaps dataset, using \( K = 3 \) layers for pruning, in line with FastV for a valid comparison.

Our innovative strategy for determining \( R \) was as follows:
\[
C_{\text{retain}} = 1 - H(L_{\text{index}} - 4) \cdot P_{\text{prune\_4th}} - H(L_{\text{index}} - 4) \cdot R'
\]
Here, \( C_{\text{retain}} \) denoted the proportion of visual tokens kept in the current layer, \( L_{\text{index}} \) represented the current layer index, \( P_{\text{prune\_4th}} \) was the fourth layer's pruning ratio, and \( R' \) was a modified pruning ratio adapting to the layer index.

The results showed that our method efficiently met Floating Point Operations Per Second (FLOPs) resource constraints without manual search, indicating the effectiveness of the adaptive \( R \).

\subsubsection{Decoding Parameters}

We also investigated how decoding strategies impacted VLM generation. Adjusting a single decoding parameter in LLaVA-1.5 while fixing others revealed the significant effect of decoding parameters on the quality of generated text. Particularly, setting top\_k to 6 resulted in the best performance, with a CIDEr score of `109.57'. To maintain fairness, we used LLaVA-1.5 with the default greedy search for replication, thereby reducing the influence of other parameters.

\subsection{Visualization}

To comprehensively elucidate the functioning of our proposed method, we visualized the alterations in image masking throughout the generation process in Figure~\ref{fig:viz2}. We employed a diverse collection of original images and introduced them to the LLaVA-1.5-7B model with the prompt: 'Provide a one-sentence caption for the provided image.' As the generation iterations escalate, we document an associated increase in the cropping ratio. This trend suggests that the model incrementally diminishes its attention on redundant information within the image during the generation, honing in on the extraction and processing of pivotal information. Additionally, by contrasting the generation trajectories of different images, we discern that our method exhibits flexibility in adapting to images of assorted types and complexities, reinforcing the efficacy and robustness of our proposed approach.

\section{Conclusion}
This paper introduces a novel method, \ourmethodName{}, to dynamically determine compression rate $R$ for VLMs during the generation process. Our approach leverages a lightweight predictor that utilizes the attention distribution across different token types to identify the most effective compression rate, thereby addressing the computational inefficiencies associated with fixed compression rates. We employ Gumbel-Softmax to overcome the non-differentiability challenges associated with traditional pruning methods.  
The empirical validation across multiple benchmarks underscores the effectiveness of our method in maintaining accuracy while reducing computational demands. This study not only enhances the efficiency of VLMs, but also paves the way for more adaptable and resource-aware applications in complex multimodal environments.

\bibliographystyle{IEEEbib}
\bibliography{main}

\end{document}